\pdfoutput=1
\documentclass[11pt]{article}

\usepackage{naacl2021} % CHANGE IT BACK TO REVIEW

\usepackage{times}
\usepackage{latexsym}

\usepackage{comment}
\usepackage{bm}
\usepackage[utf8]{inputenc}
\usepackage{amssymb}
\usepackage{amsmath}
\usepackage{multirow}
\usepackage{graphicx}
\usepackage{booktabs}
\usepackage[colorinlistoftodos]{todonotes}
\usepackage[T1]{fontenc}
\usepackage[utf8]{inputenc}

\usepackage{microtype}

\newcommand{\varuncomment}[1]{}
\newcommand{\mohitcomment}[1]{}
\newcommand{\hiroshicomment}[1]{}

\title{TABBIE: Pretrained Representations of Tabular Data}

\newcommand{\bvec}[1]{\boldsymbol{#1}}
\newcommand{\name}[0]{\textsc{tabbie}}
\newcommand{\freq}[0]{\textbf{FREQ}}
\newcommand{\mix}[0]{\textbf{MIX}}

\author{Hiroshi Iida\textsuperscript{\textnormal{\ensuremath{\dagger}}}\hspace{0.5cm}
Dung Thai\textsuperscript{\textnormal{\ensuremath{\ddagger}}}\hspace{0.5cm}
Varun Manjunatha\textsuperscript{\textnormal{\ensuremath{\S}}}\hspace{0.5cm}
Mohit Iyyer\textsuperscript{\textnormal{\ensuremath{\ddagger}}} \\
\textsuperscript{\ensuremath{\dagger}}Sony Corporation\hspace{0.5cm}
\textsuperscript{\ensuremath{\ddagger}}UMass Amherst\hspace{0.5cm}
\textsuperscript{\S}Adobe Research \\
\texttt{\normalsize hiroshi.iida@sony.com} \\
\texttt{\normalsize\{dthai,miyyer\}@cs.umass.edu} \\
\texttt{\normalsize vmanjuna@adobe.com}
}

\begin{document}
\maketitle

\begin{abstract}

Existing work on tabular representation-learning \emph{jointly} models tables and associated text using self-supervised objective functions derived from pretrained language models such as BERT. While this joint pretraining improves tasks involving paired tables and text (e.g., answering questions about tables), we show that it underperforms on tasks that operate over tables without any associated text (e.g., populating missing cells). We devise a simple pretraining objective (\emph{corrupt cell detection}) that learns exclusively from tabular data and reaches the state-of-the-art on a suite of table-based prediction tasks. Unlike competing approaches, our model (\name) provides embeddings of all table substructures (cells, rows, and columns), and it also requires far less compute to train. A qualitative analysis of our model's learned cell, column, and row representations shows that it understands complex table semantics and numerical trends. 
\end{abstract}

% The semantic understanding of tabular data is an important practical problem. However, existing pre-training methods for tabular data require many computational resources and need text-table pairs. In this paper, we propose a simple and efficient method to obtain rows, columns and cells embeddings of a table by using a new architecture with a 2d-transformer for pre-training from table data only. Our method uses lower computational resources for pre-training tables compared with existing methods. In addition, our method uses fake cell detection as a pre-training task, and we found that using a more difficult pre-training task improves the results of some downstream tasks. Experimental results show that our pre-trained model achieved equal or better results for four table-related downstream tasks compared with existing methods.

\section{Introduction}
\label{sec:introduction}

Large-scale self-supervised pretraining has substantially advanced the state-of-the-art in natural language processing~\citep{Peters:2018,devlin2018bert,Liu2019RoBERTaAR}. More recently, these pretraining methods have been extended to jointly learn representations of \emph{tables} as well as text~\citep{Herzig2020TAPASWS,yin20acl}, which enables improved modeling of tasks such as question answering over tables. However, many practical problems involve semantic understanding of tabular data without additional text-based input, such as extracting tables from documents, retrieving similar columns or cells, and filling in missing information~\cite{10.1145/3372117}. In this work, we design a pretraining methodology specifically for tables (\textbf{Tab}ular \textbf{I}nformation \textbf{E}mbedding or \name) that resembles several core tasks in table extraction and decomposition pipelines and allows easy access to representations for different tabular substructures (cells, rows, and columns).

Existing table representation models such as TaBERT~\citep{yin20acl} and TaPas~\citep{Herzig2020TAPASWS} concatenate tabular data with an associated piece of text and then use BERT's masked language modeling objective for pretraining. These approaches are computationally expensive due to the long sequences that arise from concatenating text with linearized tables, which necessitates truncating the input sequences\footnote{~\citet{Herzig2020TAPASWS} use a fixed limit of 128 tokens for both text and table, while~\citet{yin20acl} drop all but three rows of the table during pretraining.} to make training feasible. We show that TaBERT underperforms on downstream table-based applications that operate independent of external text (e.g., deciding whether cell text was corrupted while extracting a table from a PDF), which motivates us to investigate an approach that preserves the full table during pretraining.

\makeatother
\begin{figure}[t]
    \centering
  \includegraphics[width=1.0\linewidth]{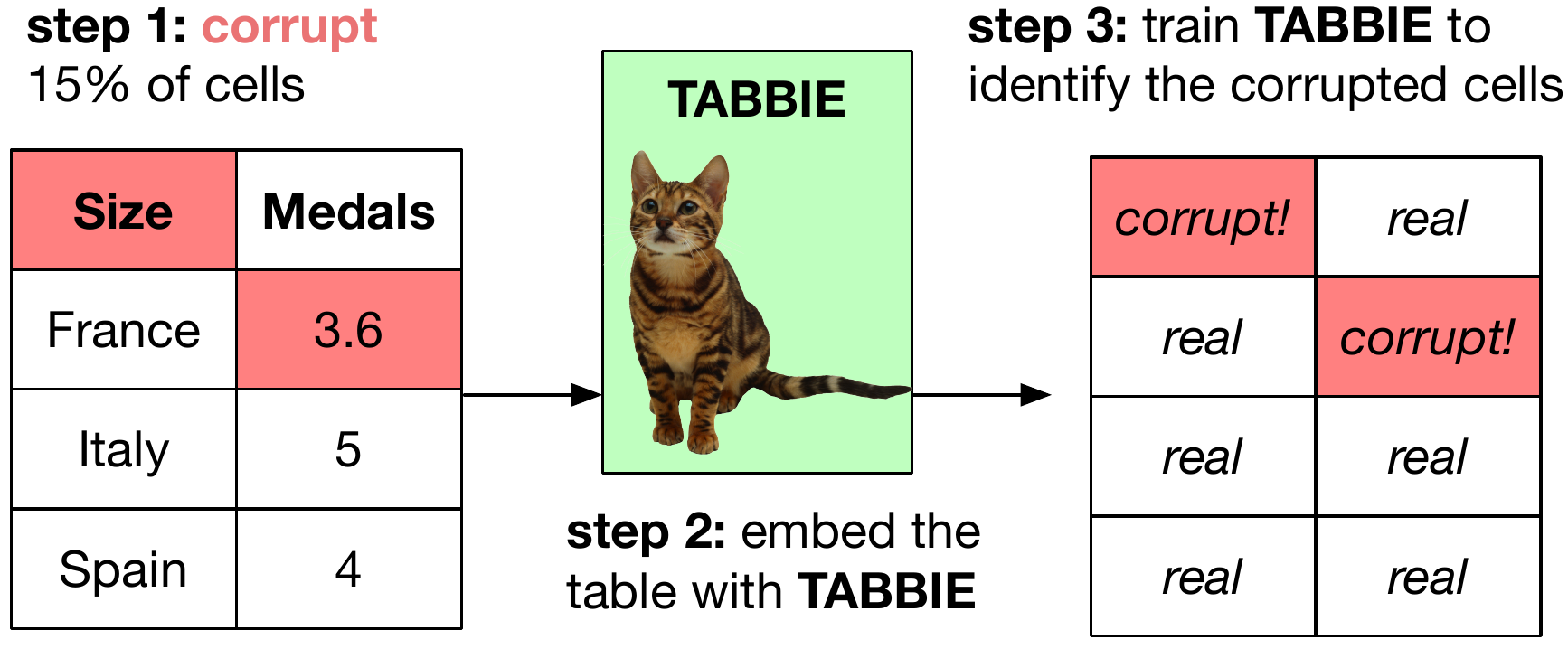}
  \caption{\name\ is a table embedding model trained to detect corrupted cells, inspired by the ELECTRA~\citep{Clark2020ELECTRA:} objective function. This simple pretraining objective results in powerful embeddings of cells, columns, and rows, and it yields state-of-the-art results on downstream table-based tasks.}
  \label{fig:overview}
\end{figure}

Our \name\ architecture relies on two Transformers that independently encode rows and columns, respectively; their representations are pooled at each layer. This setup reduces the sequence length of each Transformer's input, which cuts down on its complexity, while also allowing us to easily extract representations of cells, rows, and columns. Additionally,
\name\ uses a simplified training objective compared to masked language modeling: instead of predicting masked cells, we repurpose ELECTRA's objective function~\citep{Clark2020ELECTRA:} for tabular pretraining by asking the model to predict whether or not each cell in a table is real or corrupted. We emphasize that this pretraining objective is a fundamental task in table structure decomposition pipelines~\citep{nishida2017understanding, tensmeyer-table, raja-table}, in which incorrectly predicting row/column separators or cell boundaries leads to corrupted cell text. Unlike~\citet{Clark2020ELECTRA:}, we do not require a separate ``generator'' model that produces corrupted candidates, as we observe that simple corruption processes (e.g., sampling cells from other tables, swapping cells within the same column) yield powerful representations after pretraining. 

In a controlled comparison to TaBERT (pretraining on the same number of tables and using a similarly-sized model), we evaluate \name\ on three table-based benchmarks: column population, row population, and column type prediction. On most configurations of these tasks, \name\ achieves state-of-the-art performance, outperforming TaBERT and other baselines, while in others it performs competitively with TaBERT. Additionally, \name\ was trained on \textbf{8} V100 GPUs in just over a week, compared to the \textbf{128} V100 GPUs used to train TaBERT in six days. A qualitative nearest-neighbor analysis of embeddings derived from \name\ confirms that it encodes complex semantic properties about textual and numeric cells and substructures. We release our pretrained models and code to support further advances on table-based tasks.\footnote{\url{https://github.com/SFIG611/tabbie}}

%\todo{H: six days * 128 V100(32GB) is for TaBERT Large model. training time for TaBERT base model is not founded. our models are 14-15 days for 8 V100(16GB, mix: 14days, freq: 15days)}

% The contributions of this paper are the following : 

% \begin{enumerate}
% \item We propose an efficient method for pre-training from table data only, without the need for any additional text input. We can also obtain embeddings (i.e., features) for each row, column and cell.
% \item We use the replaced cell detection task rather than the masked language modeling task for pre-training. In addition to improving  pre-training efficiency and utilizing fewer computational resources, the more difficult setting of the replaced cell detection task achieves better accuracy for some downstream tasks.
% \item Experimental results show that our model achieves competitive or better performance for four table-related downstream tasks compared with the existing state-of-the-art.
% \end{enumerate}

\makeatother
\begin{figure*}[t]
 \centering
  \includegraphics[width=1.0\linewidth]{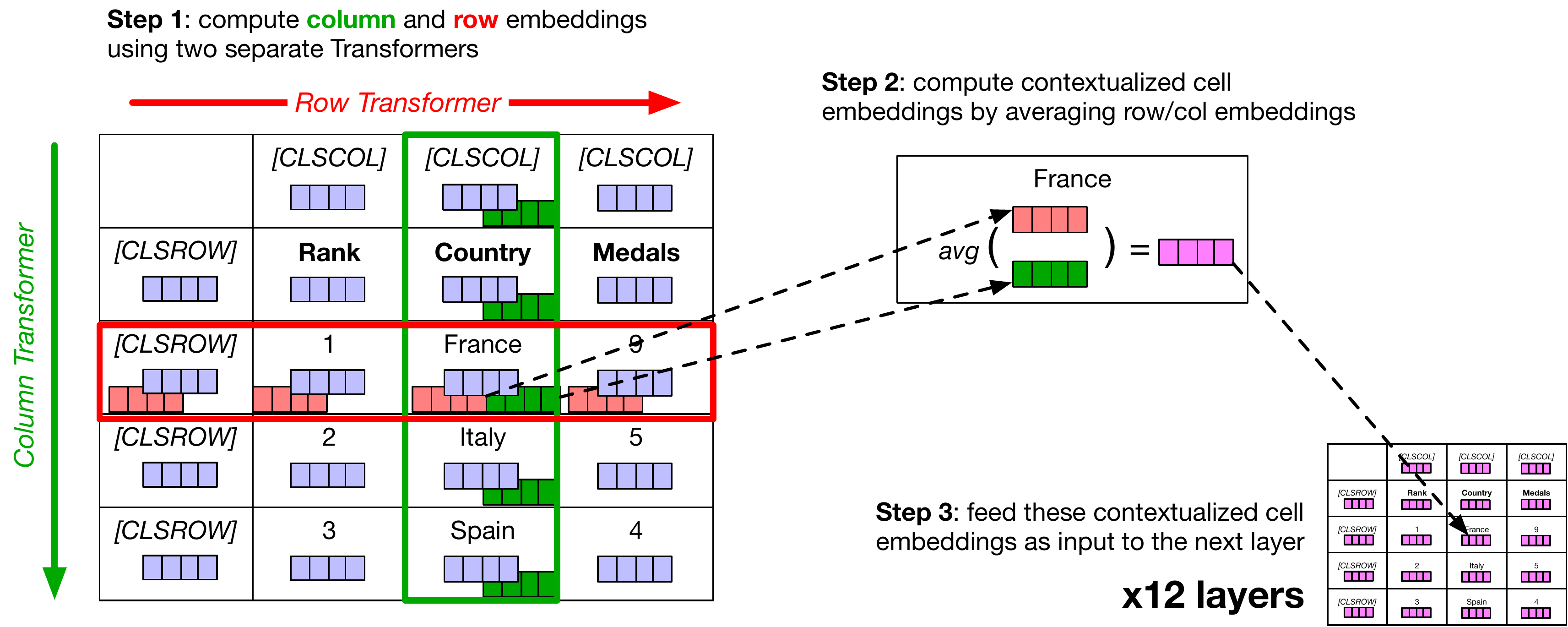}
  \caption{\name's computations at one layer. For a given table, the row Transformer contextualizes the representations of the cells in each row, while the column Transformer similarly contextualizes cells in each column. The final cell representation is an average of the row and column embeddings, which is passed as input to the next layer. \textsc{[cls]} tokens are prepended to each row and column to facilitate downstream tasks operating on table substructures.}
  \label{fig:model}
\end{figure*}

\section{Model}
\label{sec:model}

\name\ is a self-supervised pretraining approach trained exclusively on tables, unlike prior approaches~\citep{Herzig2020TAPASWS,yin20acl} that jointly model tables and associated text snippets. At a high level, \name\ encodes each cell of a table using two different Transformer models~\citep{NIPS2017_7181}, one operating across the rows of the table and the other across columns. At each layer, the representations from the \emph{row} and \emph{column} Transformers are averaged and then passed as input to the next layer, which produces a contextualized representation of each cell within the table. We place a binary classifier over \name's final-layer cell representations to predict whether or not it has been \emph{corrupted}, or replaced by an intruder cell during preprocessing, inspired by the ELECTRA objective of~\citet{Clark2020ELECTRA:}. In the remainder of this section, we formalize both \name's model architecture and pretraining objective.

\newcommand*{\Scale}[2][4]{\scalebox{#1}{$#2$}}%

\subsection{Model Architecture} 

\name\ takes an $M \times N$ table as input and produces  embeddings $\bvec{x}_{ij}$ for each cell (where $i$ and $j$ are row and column indices, respectively), as well as embeddings for individual columns $\bvec{c}_{j}$ and rows $\bvec{r}_{i}$. 

\paragraph{Initialization:} We begin by initializing the cell embeddings $\bvec{x}_{ij}$ using a pretrained BERT model~\citep{devlin2018bert}.\footnote{We use the BERT-base-uncased model in all experiments.} Specifically, for each cell $(i, j)$, we feed its contents into BERT and extract the 768-$d$ \textsc{[cls]} token representation. This step allows us to leverage the powerful semantic text encoder of BERT to compute representations of cells out-of-context, which is important because many tables contain cells with long-form text (e.g., \emph{Notes} columns). Additionally, BERT has been shown to encode some degree of numeracy~\citep{Wallace2019Numeracy}, which helps represent cells with numerical content. We keep this BERT encoder fixed during training to reduce computational expense. Finally, we add learned positional embeddings to each of the \textsc{[cls]} vectors to form the initialization of $\bvec{x}_{ij}$. More specifically, we have two sets of positional embeddings, $p^{(r)}_i \in \mathbb{R}^H$ and $p^{(c)}_j \in \mathbb{R}^H$, which model the position of rows and columns, respectively, and are randomly initialized and fine-tuned via \name's self-supervised objective.

\paragraph{Contextualizing the cell embeddings:}
The cell embeddings we get from BERT are uncontextualized: they are computed in isolation of all of the other cells in the table. While methods such as TaBERT and TaPaS contextualize cell embeddings by linearizing the table into a single long sequence, we take a different and more computationally manageable approach. We define a \emph{row} Transformer, which encodes cells across each row of the table, as well as a \emph{column} Transformer, which does the same across columns. 

Concretely, assume row $i$ contains cell embeddings $\bvec{x}_{i,1}, \bvec{x}_{i,2}, \dots, \bvec{x}_{i,N}$. We pass this sequence of embeddings into a \emph{row} Transformer block, which uses self-attention to produce contextualized output representations $\bvec{r}_{i,1}, \bvec{r}_{i,2}, \dots, \bvec{r}_{i,N}$. Similarly, assume column $j$ contains cell embeddings $\bvec{x}_{1,j}, \bvec{x}_{2,j}, \dots, \bvec{x}_{M,j}$; the \emph{column} Transformer produces contextualized representations $\bvec{c}_{1,j}, \bvec{c}_{2,j}, \dots, \bvec{c}_{M,j}$. After running the two Transformers over all rows and columns, respectively, each cell $(i, j)$ of a table is associated with a row embedding $\bvec{r}_{i,j}$ as well as a column embedding $\bvec{c}_{i,j}$. 

The final step of cell contextualization is to compose the row and column embeddings together before feeding the result to the next layer. Intuitively, if we do not aggregate the two sets of embeddings together, subsequent layers of the model will only have access to information from a specific row or column, which prevents contextualization across the whole table. We implement this aggregation through simple averaging: specifically, at layer $L$ of \name, we compute cell embeddings as:
\begin{equation}
    \bvec{x}^{L+1}_{i,j} = \frac{\bvec{r}^L_{i,j} + \bvec{c}^L_{i,j}}{2}
\end{equation}

The new cell representations $\bvec{x}^{L+1}_{i,j}$ are then fed to the row and column Transformers at the next layer $L+1$. 

\paragraph{Extracting representations of an entire row or column:}
The row and column Transformers defined above produce separate representations for every cell in a particular row or column. However, many table-related downstream tasks (e.g., retrieve similar columns from a huge dataset of tables to some query column) can benefit from embeddings that capture the contents of an entire row or column. To enable this functionality in \name, we simply prepend \textsc{[clsrow]} and \textsc{[clscol]} tokens to the beginning of each row and column in an input table as a preprocessing step. After pretraining, we can extract the final-layer cell representations of these  \textsc{[cls]} tokens to use in downstream tasks. 

\subsection{Pretraining}
Having described \name's model architecture, we turn now to its training objective. We adapt the self-supervised ELECTRA objective proposed by~\citet{Clark2020ELECTRA:} for text representation learning, which places a binary classifier over each word in a piece of text and asks if the word either is part of the original text or has been corrupted. While this objective was originally motivated as enabling more efficient training compared to BERT's masked language modeling objective, it is especially suited for tabular data, as corrupt cell detection is actually a fundamental task in table structure decomposition pipelines such as \cite{nishida2017understanding, tensmeyer-table, raja-table}, in which incorrectly predicted row/column separators or cell boundaries can lead to corrupted cell text.

In our extension of ELECTRA to tables, a binary classifier takes a final-layer cell embedding as input to decide whether it has been corrupted. More concretely, for cell $(i, j)$, we compute the corruption probability as 

\begin{equation}
    P_{\text{corrupt}}(\text{cell}_{i, j}) = \sigma(\bvec{w}^\intercal \bvec{x}^L_{i,j})
\end{equation}

where $L$ indexes \name's final layer, $\sigma$ is the sigmoid function, and $\bvec{w}$ is a weight vector of the same dimensionality as the cell embedding. Our final loss function is the binary cross entropy loss of this classifier averaged across all cells in the table. 

% \todo{H: The total loss is an average of the (the average header loss) and the (the average cell loss)}

\subsection{Cell corruption process}
\label{subsec:corruption}
\begin{figure}
 %\begin{flushleft}
    \centering
  % ¥includegraphics[width=0.3¥textwidth,natwidth=500,natheight=500]{fake_tables.png}
  \includegraphics[width=1.0\linewidth]{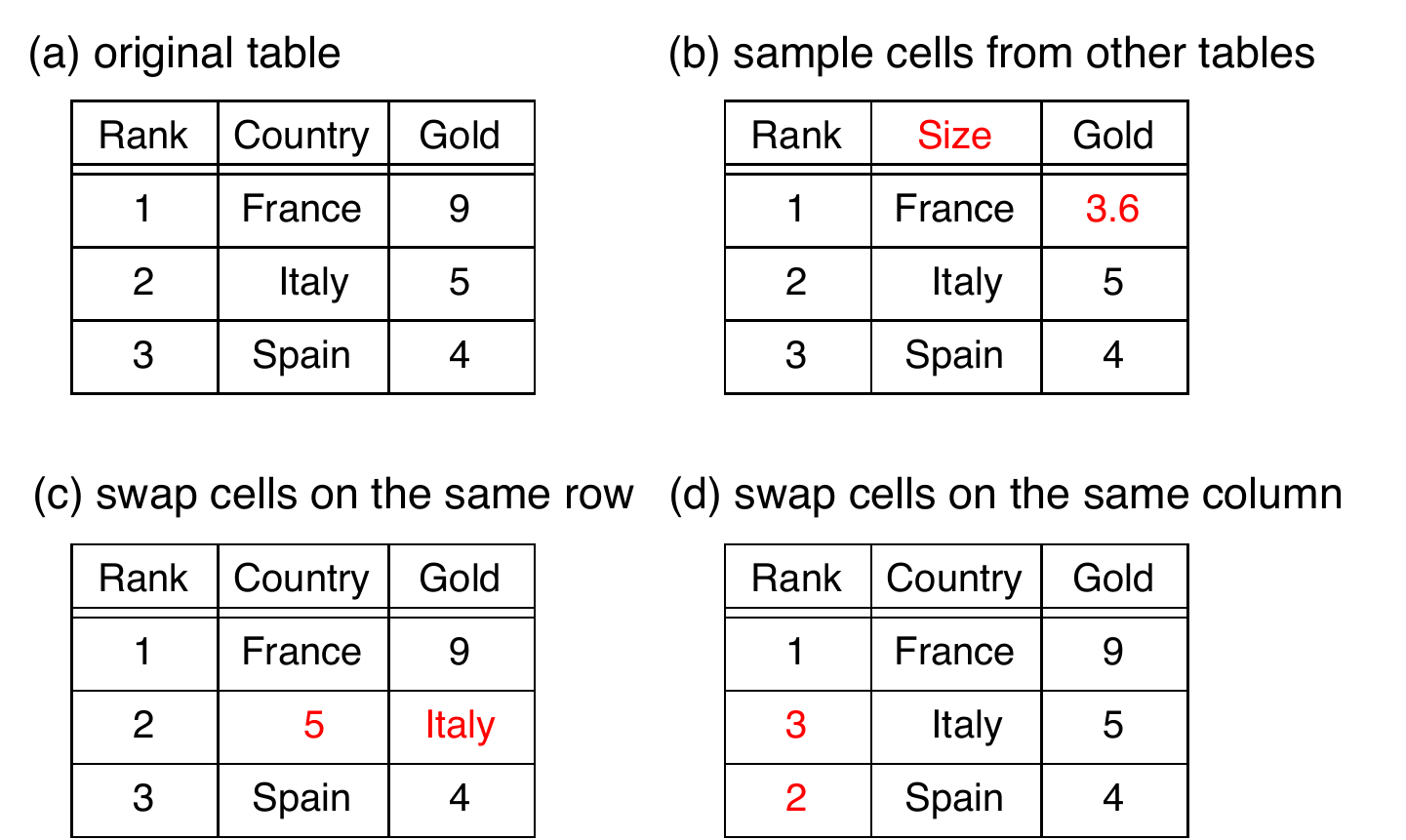}
  \caption{The different cell corruption strategies used in our experiments.}
  \label{fig:fake_tables}
 %\end{flushleft}
\end{figure}

Our formulation diverges from \citet{Clark2020ELECTRA:} in how the corrupted cells are generated. In ELECTRA, a separate generator model is trained with BERT's masked language modeling objective to produce candidate corrupted tokens: for instance, given \emph{Jane went to the \textsc{[mask]} to check on her experiments}, the generator model might produce corrupted candidates such as \emph{lab} or \emph{office}. Simpler corruption strategies, such as randomly sampling words from the vocabulary, cannot induce powerful representations of text because local syntactic and semantic patterns are usually sufficient to detect obvious corruptions. For tabular data, however, we show that simple corruption strategies (Figure~\ref{fig:fake_tables}) that take advantage of the intra-table structure actually do yield powerful representations without the need of a separate generator network. More specifically, we use two different corruption strategies:
 
\begin{itemize} 
    \item \textbf{Frequency-based cell sampling}: Our first strategy simply samples corrupt candidates from the training cell frequency distribution (i.e., more commonly-occurring cells are sampled more often than rare cells). One drawback of this method is that oftentimes it can result in samples that violate a particular column type (for instance, sampling a textual cell as a replacement for a cell in a numeric column). Despite its limitations, our analysis in Section~\ref{sec:analysis} shows that this strategy alone results in strong performance on most downstream table-based tasks, although it does not result in a rich semantic understanding of intra-table semantics. 
    
    \item \textbf{Intra-table cell swapping}: To encourage the model to learn fine-grained distinctions between topically-similar data, our second strategy produces corrupted candidates by swapping two cells in the same table (Figure~\ref{fig:fake_tables}c, d). This task is more challenging than the frequency-based sampling strategy above, especially when the swapped cells occur within the same column. While it underperforms frequency-based sampling on downstream tasks, it qualitatively results in more semantic similarity among nearest neighbors of column and row embeddings.
\end{itemize}

\makeatother

\subsection{Pretraining details}

\noindent\textbf{Data:}
We aim for as controlled of a comparison with TaBERT~\citep{yin20acl} as possible, as its performance on table QA tasks indicate the strength of its table encoder. TaBERT's pretraining data was not publicly released at the time of our work, but their dataset consists of 26.6M tables from Wikipedia and the Common Crawl. We thus form a pretraining dataset of equivalent size by combining 1.8M Wikipedia tables with 24.8M preprocessed Common Crawl tables from Viznet \cite{DBLP:conf/chi/HuGHBZHK0SD19}.\footnote{The vast majority of text in these tables is in English.} 

\paragraph{Experimental settings:} We train \name\ for seven epochs for just over a week on 8 V100 GPUs using mixed precision. \name\ has 12 layers and a hidden dimensionality of $768$ for both row and column Transformers, in an effort to be comparably-sized to the TaBERT-Base model.\footnote{\name\ is slightly larger than TaBERT-Base (170M to 133M parameters) because its row and column Transformers are the same size, while TaBERT places a smaller ``vertical'' Transformer over the output of a fine-tuned BERT model.}  Before computing the initial cell embeddings using BERT, we truncate each cell's contents to the first 300 characters, as some cells contain huge amounts of text. We also truncate tables to 30 rows and 20 columns to avoid memory issues (note that this is much larger than the three rows used by TaBERT), and our maximum batch size is set at 4,800 cells (on average, 104 tables per batch). We use the Adam optimizer~\citep{kingma2014adam} with a learning rate of 1e-5.

We compared two pretrained models trained with different cell corruption strategy for downstream tasks. The first strategy (\freq) uses exclusively a frequency-based cell sampling. The second strategy is a 50/50 mixture (\mix) of frequency-based sampling and intra-table cell swapping, where we additionally specify that half of the intra-table swaps must come from the same row or column to make the objective more challenging.

\section{Experiments}
\label{sec:experiments}

We validate \name's table representation quality through its performance on three downstream table-centric benchmarks (column population, row population, and column type prediction) that measure semantic table understanding. In most configurations of these tasks, \name\ outperforms TaBERT and other baselines to set new state-of-the-art numbers. Note that we do \emph{not} investigate \name's performance on table-and-text tasks such as WikiTableQuestions~\citep{pasupat2015compositional}, as our focus is not on integrating \name\ into complex task-specific pipelines~\citep{liang2018memory}, although this is an interesting avenue for future work.

% \subsection{experimental setup}

\makeatother
\begin{figure}
    \centering
 %\begin{flushleft}
  % ¥includegraphics[width=0.3¥textwidth,natwidth=500,natheight=500]{fake_tables.png}
  \includegraphics[width=1.0\linewidth]{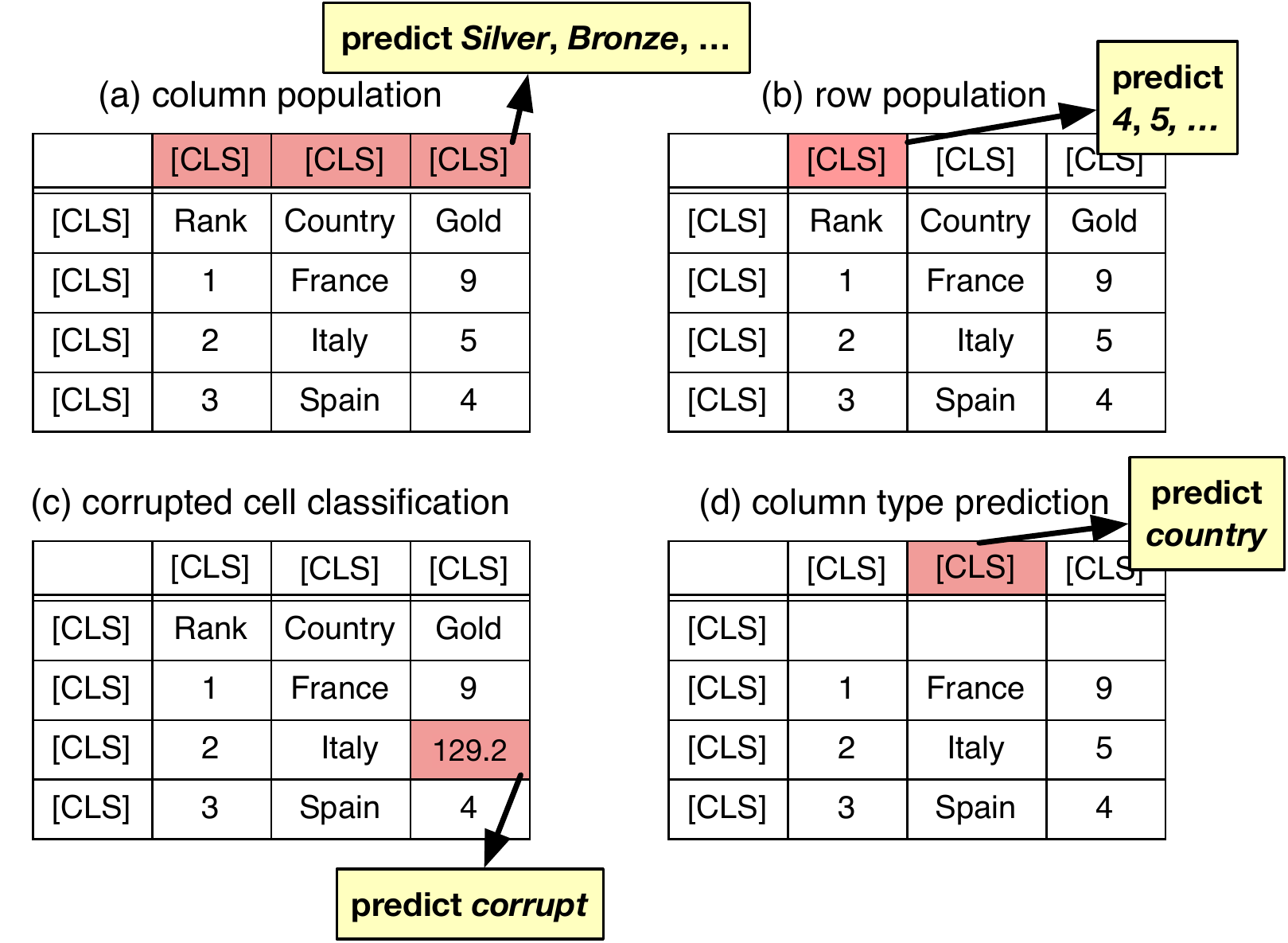}
  \caption{The inputs and outputs for each of our table-based prediction tasks. Column type prediction does not include headers as part of the table.}
  \label{fig:finetune}
 %\end{flushleft}
\end{figure}
\begin{table}
\centering
\scalebox{0.8}{
\begin{tabular}{lccc}
\toprule
\textbf{Task} & \textbf{Batch size} & \textbf{LR} & \textbf{Max epochs} \\
\midrule
Column population & {12} & {1e-05} & {20} \\
Row population & {48} & {2e-05}  & {30} \\
% corrupt cell detection & {1} & {0.2} & {10} \\
Col. type prediction & {12} & {2e-05} & {15} \\
\bottomrule
\end{tabular}}
\caption{Fine-tuning hyperparameters of each downstream task for \name\ and TaBERT.}\label{tab:params_finetune}
\end{table}

\subsection{Fine-tuning \name}
In all of our downstream experiments, we apply essentially the same fine-tuning strategy to both \name\ and TaBERT: we select a subset of its final-layer representations (i.e., cell or column representations) that correspond to the tabular substructures used in the downstream task, and we place a classifier over these representations to predict the training labels. We select task-specific hyperparameters based on the size of each dataset (full details in Table~\ref{tab:params_finetune}) and report the test performance of the best-performing validation checkpoint. For both models, we backpropagate the downstream error signal into all of the model's parameters (i.e., we do not ``freeze'' our pretrained model).

\subsection{Column Population}
% \paragraph{problem setting:}
In the column population task, which is useful for attribute discovery, tabular data augmentation, and table retrieval \cite{dassarma-finding}, a model is given the first $N$ columns of a ``seed'' table and asked to predict the remaining column headers.~\citet{entitables} compile a dataset for this task comprising 1.6M tables from Wikipedia with a test set of 1,000 tables, formulated as a multi-label classification task with 127,656 possible header labels. Importantly, we remove all of the tables in the column population test set from our pretraining data to avoid inflating our results in case \name\ memorizes the missing columns during pretraining.\footnote{Note that TaBERT's pretraining data likely includes the test set tables, which may give it an advantage in our comparisons.} 

To fine-tune \name\ on this task, we first concatenate the column \textsc{[clscol]} embeddings of the seed table into a single vector and pass it through a single linear and softmax layer, training with a multi-label classification objective~\cite{Mahajan2018ExploringTL}. Our baselines include the generative probabilistic model (\textbf{GPM}) of~\citet{entitables} as well as a word embedding-based extension called Table2VecH (\textbf{TH}) devised by~\citet{Zhang:2018:AHT}. As fine-tuning on the full dataset is extremely expensive for \name\ and TaBERT, we fine-tune on a random subset of 100K training examples; as a further disadvantage to these, we do not use table captions (unlike GPM and GPM+TH) during training. Nevertheless, as Table~\ref{tab:exp_cp} shows, \name\ and TaBERT substantially outperform both baselines, and \name\ consistently outperforms TaBERT regardless of how many seed columns are provided, especially with only one seed column. This result indicates that \name\ encodes more semantics about headers and columns than TaBERT.

\begin{table}
\centering
\scalebox{0.78}{
\begin{tabular}{lcccccccc}
\toprule
N & Method & MAP & MRR & Ndcg-10 & Ndcg-20 \\
\midrule
\multirow{4}{*}{ 1 }  & GPM & {25.1} & {37.5} & {-}& {-}  \\
& GPM+TH & {25.5} & {0.38.0} & {27.1} & {31.5}  \\
& TaBERT & {33.1} & {41.3}  & {35.1} & {38.1} \\ 
& \name\ (FREQ) & {\textbf{37.9}} & {\textbf{49.1}}  & {\textbf{41.2}} & {\textbf{43.8}} \\ 
& \name\ (MIX) & {37.1} & {48.7}  & {40.4} & {43.1} \\ \midrule
\multirow{4}{*}{  2 }  & GPM & {28.5} & {40.4} & {-}& {-}  \\
& GPM+TH & {33.2} & {44.0} & {36.1} & {41.3}  \\
& TaBERT & {51.1} & {60.1}  & {54.7} & {56.6} \\
& \name\ (FREQ) & {\textbf{52.0}} & {\textbf{62.8}}  & {\textbf{55.8}} & {\textbf{57.6}} \\ 
& \name\ (MIX) & {51.7} & {62.3}  & {55.6} & {57.2} \\ \midrule
\multirow{4}{*}{ 3 }  & GPM & {28.5} & {35.5} & {-}& {-}  \\
& GPM+TH & {40.0} & {50.8} & {45.2} & {48.5}  \\
& TaBERT & {53.3} & {60.9}  & {56.9} & {57.9} \\
& \name\ (FREQ) & {\textbf{54.5}} & {\textbf{63.3}}  & {\textbf{57.9}} & {\textbf{58.9}} \\ 
& \name\ (MIX) & {54.1} & {62.3}  & {57.4} & {58.7} \\ \bottomrule

\end{tabular}%
}
%}
\caption{\name\ outperforms all methods on the column population task, with the biggest improvement coming with just a single seed column  ($N=1$). Despite its simplicity, the \freq\ corruption strategy yields better results than \mix.}\label{tab:exp_cp}
\end{table}

\subsection{Row Population}
The row population task is more challenging than column population: given the first $N$ rows of a table in which the first column contains entities (e.g., ``Country''), models must predict the remaining entries of the first column. Making reasonable predictions of which entities best fill the column requires understanding the full context of the seed table.
The~\citet{entitables} dataset also contains a split for row population, which we use to evaluate our models. Again, since the dataset is too large for our large embedding models, we sample a subset of tables for fine-tuning.\footnote{We sample all tables that have at least five entries in the left-most column, which results in roughly 200K tables.} Our label space consists of 300K entities that occur at least twice in Wikipedia tables, and we again formulate this problem as multi-label classification, this time on top of the first column's \textsc{[clscol]} representation.\footnote{Due to the large number of labels, we resort to negative sampling during training instead of the full softmax to cut down on fine-tuning time. Negative samples are formed by uniform random sampling on the label space.} 

On this task, TaBERT and \name\ again outperform the baseline Entitables model (which uses external information in the form of table captions). When given only one seed row, TaBERT slightly outperforms \name, but with more seed rows, \name\ exhibits small improvements over TaBERT. 

% Observing results from Table \ref{tab:exp_rp}, we note that though TaBERT scores higher for single row tables ($N=1$), \name\ gets better results for $N=2$ and $3$, i.e., \name\ is better at predicting remaining cells given larger input tables. Both of these methods significantly outperform the Entitables \cite{entitables} baseline, which utilizes table captions.

\begin{table}
\centering
\scalebox{0.78}{
\begin{tabular}{lcccccccc}
\toprule
N & Method & MAP & MRR & Ndcg-10 & Ndcg-20 \\
\midrule
\multirow{4}{*}{ 1 }  & Entitables & {36.8} & {45.2}  & {-} & {-} \\
& TaBERT & {\textbf{43.2}} & {\textbf{55.7}}  & {\textbf{45.6}} & {\textbf{47.7}} \\ 
& \name\ (FREQ) & {42.8} & {54.2}  & {44.8} & {46.9} \\ 
& \name\ (MIX) & {42.6} & {54.7}  & {45.1} & {46.8} \\ \midrule
\multirow{4}{*}{  2 }  & Entitables & {37.2} & {45.1}  & {-} & {-} \\
& TaBERT & {43.8} & {56.0}  & {46.4} & {48.8} \\ 
& \name\ (FREQ) & {\textbf{44.4}} & {\textbf{57.2}}  & {\textbf{47.1}} & {\textbf{49.5}} \\ 
& \name\ (MIX) & {43.7} & {55.7}  & {46.2} & {48.6} \\ \midrule
\multirow{4}{*}{ 3 }  & Entitables & {37.1} & {44.6}  & {-} & {-} \\
& TaBERT & {42.9} & {55.1}  & {45.6} & {48.5} \\ 
& \name\ (FREQ) & {\textbf{43.4}} & {\textbf{56.5}}  & {\textbf{46.6}} & {\textbf{49.0}} \\ 
& \name\ (MIX) & {42.9} & {55.5}  & {45.9} & {48.3} \\ \bottomrule

\end{tabular}%
}
%}
\caption{\name\ outperforms baselines on row population when provided with more seed rows $N$, although TaBERT is superior given just a single seed row. Again, the \freq\ strategy produces better results than \mix.}\label{tab:exp_rp}
\end{table}

\subsection{Column Type Prediction}
While the prior two tasks involve predicting missing elements of a table, the column type prediction task involves predicting a high-level \emph{type} of a particular column (e.g., \emph{name}, \emph{age}, etc.) without access to its header. This task is useful when indexing tables with missing column names, which happens relatively often in practice, or for schema matching\cite{Hulsebos2019SherlockAD, journals/vldb/RahmB01}, and like the other tasks, requires understanding the surrounding context. We evaluate our models on the same subset of VizNet Web Tables~\citep{DBLP:conf/chi/HuGHBZHK0SD19}\footnote{Again, we ensure that none of the test tables in this dataset occur in \name's pretraining data.} created by~\citet{zhang2019sato} to evaluate their column type predictor, SATO\footnote{\url{https://github.com/megagonlabs/sato}}. They formulate this task as a multi-class classification problem (with 78 classes), with a training set of 64,000 tables and a test set consisting of 16,000 tables. We set aside 6,400 training tables to form a validation for both \name\ and TaBERT, and we fine-tune each of these models with small random subsets of the training data (1000 and 10000 labeled tables) in addition to the full training set to evaluate their performance in a simulated low-resource setting. 

% \todo{500 data: tabbie(80.9/79.0), tabert(80.6/79.6), 1000 data: tabbie(85.6/84.7), tabert(85.2/84.7), 5000 data: currently runnning}

% To fine-tune \name\ on this task, we add a single linear layer and softmax to \textsc{[CLS]} token embeddings corresponding to the column to be predicted, as in previous tasks. In this problem setting, column headers are obviously not used, so we first replace them with blank headers before fine-tuning (Figure \ref{fig:finetune} (a)). 
Along with TaBERT, we compare with two recently-proposed column type prediction methods: Sherlock~\cite{Hulsebos2019SherlockAD}, which uses a multi-input neural network with hand-crafted features extracted from each column, and the aforementioned SATO~\cite{zhang2019sato}, which improves Sherlock by incorporating table context, topic model outputs, and label co-occurrence information. % (Do I add details about LDA and CRFs) 
% Sherlock\cite{DBLP:conf/chi/HuGHBZHK0SD19} and Sato\cite{zhang2019sato} are the state of the art methods without pre-training, while TaBERT and ours are the pre-trained model. 
Table \ref{tab:exp_ctp_full} shows the support-weighted F1-score for each method. Similar to the previous two tasks, \name\ and TaBERT significantly outperform the prior state-of-the-art (SATO). Here, there are no clear differences between the two models, but both reach higher F1 scores than the other baselines even when given only 1,000 training examples, which demonstrates the power of table-based pretraining.
% We notice that TaBERT and our method, \name\ achieve better results on this task, thus demonstrating the effectiveness of the pre-training. In fact, \name\ and TaBERT trained with roughly one-sixth of the data outperform Sato and Sherlock trained on the entire training set. % TaBERT and ours are the similar results for the column type prediction. Table. We also compared the different number of training data. When we use  $n=10,000$ training data, the result is better than no-pretraining methods \cite{DBLP:conf/chi/HuGHBZHK0SD19} and \cite{zhang2019sato},   

% \input{tables/exp_ctp_full}

\begin{table}
\centering
\scalebox{0.85}{
\begin{tabular}{lccc}
\toprule
Method & $n$=1000 & $n$=10000 & $n$=all \\
\midrule
Sherlock & - & - & {86.7} \\
SATO & - & - & {90.8} \\
TaBERT & \textbf{84.7} & {93.5} & \textbf{{97.2}} \\
\name\ (FREQ) & \textbf{84.7} & {\textbf{94.2}} & {96.9} \\ 
\name\ (MIX) & {84.1} & {93.8} & {96.7} \\ \bottomrule

\end{tabular}
}
\caption{Support-weighted F1-score of different models on column type prediction. TaBERT and \name\ perform similarly in low resource settings ($n$=1000) and when the full training data is used ($n$=all).}\label{tab:exp_ctp_full}
\end{table}

\section{Analysis}
\label{sec:analysis}

The results in the previous section show that \name\ is a powerful table representation method, outperforming TaBERT in many downstream task configurations and remaining competitive in the rest. In this section, we dig deeper into \name's representations by comparing them to TaBERT across a variety of quantitative and qualitative analysis tasks, including our own pretraining task of corrupt cell classification, as well as embedding clustering and nearest neighbors. Taken as a whole, the analysis suggests that \name\ is able to better capture fine-grained table semantics. 

\subsection{Corrupt Cell Detection}
We first examine how TaBERT performs on \name's pretraining task of corrupt cell detection, which again is practically useful as a post-processing step after table structure decomposition \cite{tensmeyer-table, raja-table} because mistakes in predicting row/column/cell boundaries (sometimes compounded by OCR errors) can lead to inaccurate extraction. We fine-tune TaBERT on 100K tables using the \mix\ corruption strategy for ten epochs, and construct a test set of 10K tables that are unseen by both TaBERT and \name\ during pretraining. While \name\ of course sees an order of magnitude more tables for this task during pretraining, this is still a useful experiment to determine if TaBERT's pretraining objective enables it to easily detect corrupted cells.  

As shown in Table~\ref{tab:exp_fcd_cell}, TaBERT performs significantly worse than \name\ on all types of corrupt cells (both random corruption and intra-table swaps). Additionally, intra-column swaps are the most difficult for both models, as \name\ achieves a \textbf{68.8} F1 on this subset compared to just \textbf{23.7} F1 by TaBERT. Interestingly, while the \mix\ strategy consistently performs worse than \freq\ for the \name\ models evaluated on the three downstream tasks in the previous section, it is substantially better at detecting more challenging corruptions, and is almost equivalent to detecting random cells sampled by \freq. This result indicates that perhaps more complex table-based tasks are required to take advantage of representations derived using \mix\ corruption.

% For this task, we do not perform explicit fine-tuning, but used our pre-trained model directly, since the model can predict the error probability of each cell and header (as shown in Figure\ref{fig:finetune} (d)).
% For comparison, we use 100,000 tables for fine-tuning TaBERT for 10 epochs. We now evaluate TaBERT and \name\ on 10,000 test tables. The tables in training and test datasets are generated randomly using the following four corrupt cell generation strategies (explained earlier in Section \ref{subsec:corruption}) : 1. Frequency-based cell sampling, 2. Intra-table cell swapping, 3. Intra-row cell swapping and 4. Intra-column cell swapping. The results of the corrupt cell classification task are shown in Tables \ref{tab:exp_fcd_header} and \ref{tab:exp_fcd_cell}. We note that \name\ outperforms TaBERT in all configurations of this task. Comparing different configurations of \name, we note that cell-swapping strategy (mix) is significantly better at correctly classifying corrupted cells.

% \input{tables/exp_fcd_header}
\begin{table}
\centering
\scalebox{0.77}{
\begin{tabular}{lcccccccc}
\toprule
Corruption & Method & Prec. & Rec. & F1\\
\midrule
\multirow{2}{*}{ \emph{Intra-row swap} } & TaBERT & {85.5} & {83.0} & {84.2}  \\ 
& \name\ (FREQ) & {99.0} & {81.4}  & {89.4}  \\
& \name\ (MIX) & {99.6} & {95.8}  & {\textbf{97.7}}  \\ \midrule
\multirow{2}{*}{  \emph{Intra-column swap} } & TaBERT & {31.2} & {19.0}  & {23.7} \\ 
& \name\ (FREQ) & {90.9} & {22.3}  & {35.8} \\
& \name\ (MIX) & {91.5} & {55.0}  & {\textbf{68.8}} \\ \midrule
\multirow{2}{*}{ \emph{Intra-table swap} } & TaBERT & {81.2} & {69.5}  & {74.9} \\ 
& \name\ (FREQ) & {98.2} & {73.3}  & {84.0} \\
& \name\ (MIX) & {98.4} & {86.2}  & {\textbf{91.9}} \\ \midrule
\multirow{2}{*}{ \emph{Random FREQ cell} } & TaBERT & {86.7} & {87.0}  & {86.8} \\ 
& \name\ (FREQ) & {99.3} & {98.2}  & {\textbf{98.8}} \\
& \name\ (MIX) & {99.1} & {98.1}  & {98.6} \\ \midrule
\multirow{2}{*}{ \emph{All} } & TaBERT & {75.6} & {65.2}  & {70.0} \\ 
& \name\ (FREQ) & {98.2} & {69.5}  & {81.4} \\ 
& \name\ (MIX) & {97.8} & {84.1}  & {\textbf{90.5}} \\ \bottomrule

\end{tabular}%
}
%}
\caption{A fine-grained comparison of different models on corrupt cell detection, with different types of corruption. TaBERT struggles on this task, especially in the challenging setting of \emph{intra-column swaps}. Unlike our downstream tasks, the \mix\ strategy is far superior to \freq\ here.}\label{tab:exp_fcd_cell}
\end{table}

%\subsection{Clustering Results}

% \input{figs/name_col}
% \input{figs/nn_row}

\makeatother
\begin{figure}[t]
    \centering
 % \begin{flushleft}
  \includegraphics[width=1.0\linewidth]{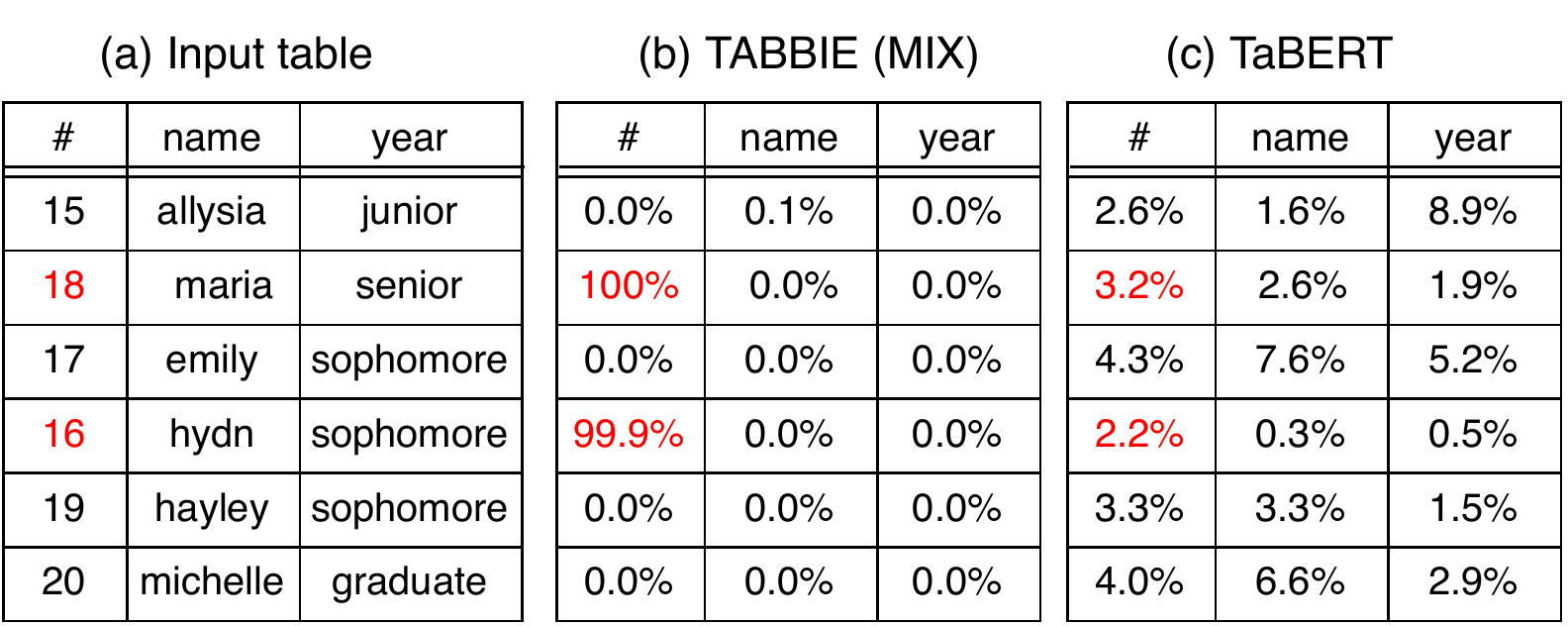}
  \caption{In this figure, (b) and (c) contain the predicted corruption probability of each cell in (a). Only \name\ \mix\ is able to reliably identify violations of numerical trends in columns. }
  \label{fig:num_understand1}
 % \end{flushleft}
\end{figure}

% \input{figs/num_understand2}

% \input{figs/nearest_neighbor}
%First, we compared the clustering result of column embeddings by applying the t-SNE. %Fig.\ref{fig:name_col} is the t-SNE result of \name\  and TaBERT for the "name" columns. We %extract the most frequent eight column combinations that include "name" column, then apply t-SNE %for the column embeddings of these methods. The same color of the dots mean their remaining %headers in these tables are the same. The left figure is the clustering result of TaBERT column %embeddings. It basically works but green dots and blue dots are not separated. The right figure %is the result of \name\ . Each color is completely separeted in this figure, this means the %column embedding of \name\ has remaining table context more. The experiment result on the column %population task also indicates the header embedding of \name\ has more information about the %remaining headers.

% \input{tables/exp_pop}

\subsection{Nearest neighbors}
\begin{figure}[t]
    \centering
  \includegraphics[width=0.9\linewidth]{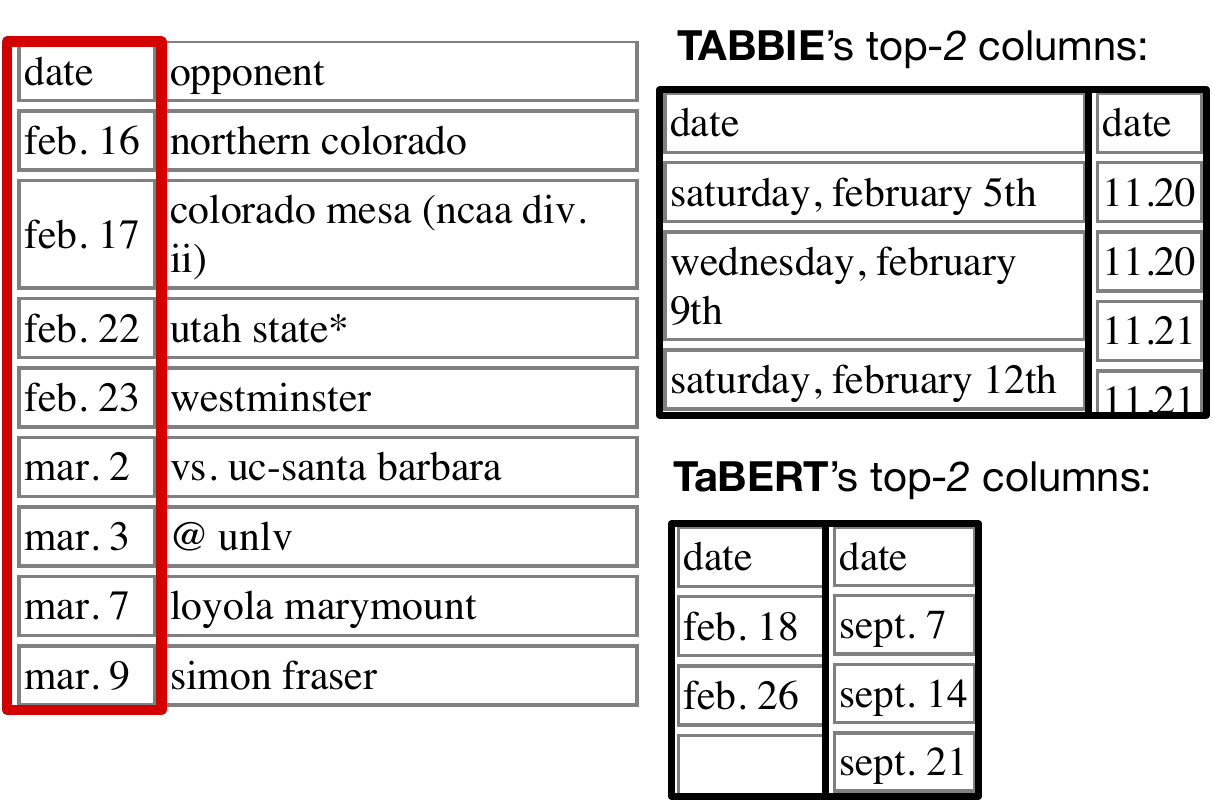}

  \bigbreak
  \includegraphics[width=0.9\linewidth]{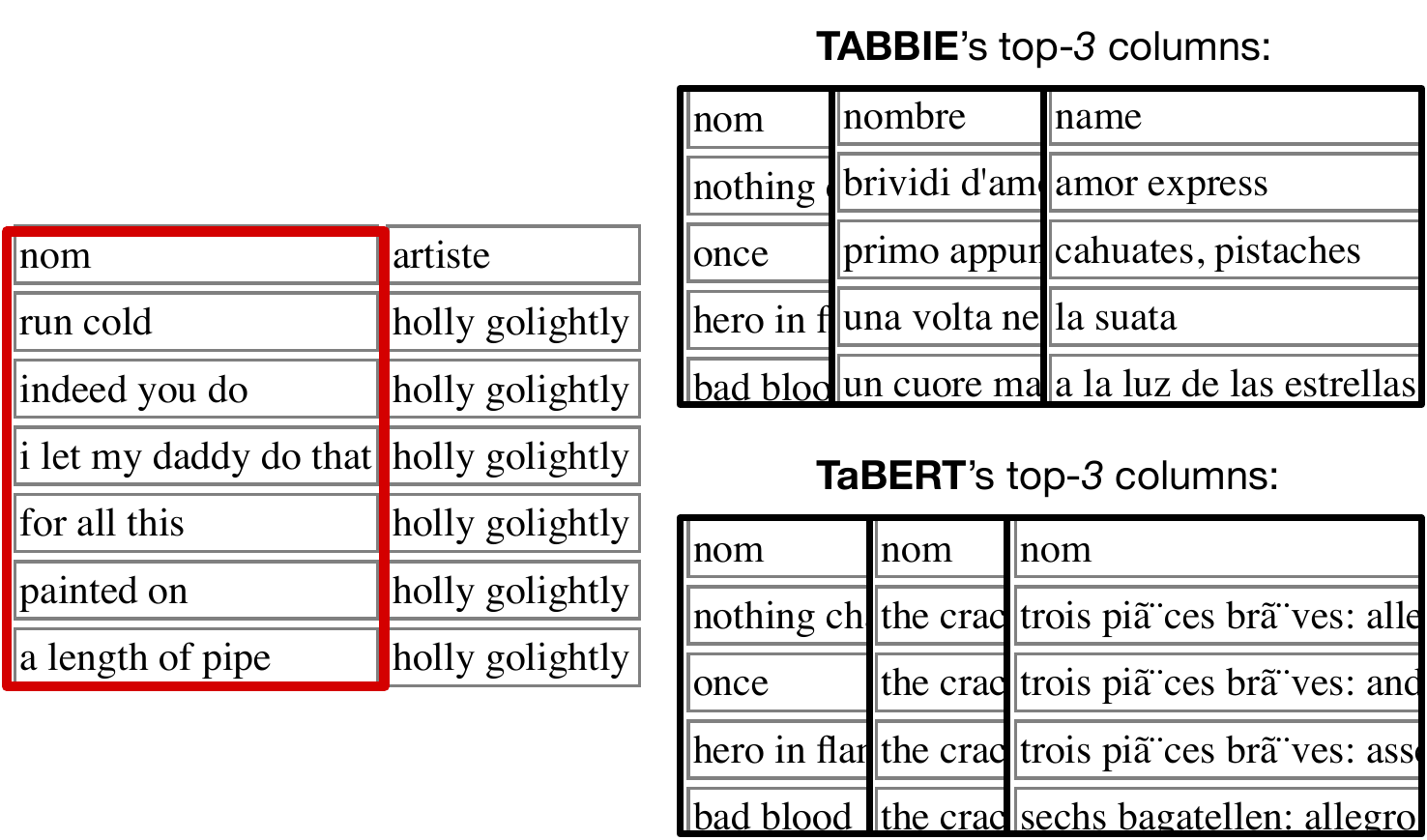}
    \caption{Nearest neighbors of the  \emph{date} and \emph{nom} columns from the tables on the left, from both \name\ and TaBERT. \name's nearest neighbors exhibit more diverse formatting and less reliance on the header, which is an example of its semantic representation capability.}
  \label{fig:date_example}
 % \end{center}
\end{figure}

We now turn to a qualitative analysis of the representations learned by \name. In Figure~\ref{fig:date_example} (top), we display the two nearest neighbor columns from our validation set to the \emph{date} column marked by the red box. \name\ is able to model the similarity of \emph{feb. 16} and \emph{saturday. february 5th} despite the formatting difference, while TaBERT's neighbors more closely resemble the original column. Figure~\ref{fig:date_example} (bottom) shows that \name's nearest neighbors are less reliant on matching headers than TaBERT, as the neighbors all have different headers (\emph{nom}, \emph{nombre}, \emph{name}). 
\makeatother

\subsection{Clustering}
\begin{figure*}[t]
 \centering
  \includegraphics[width=1.0\linewidth]{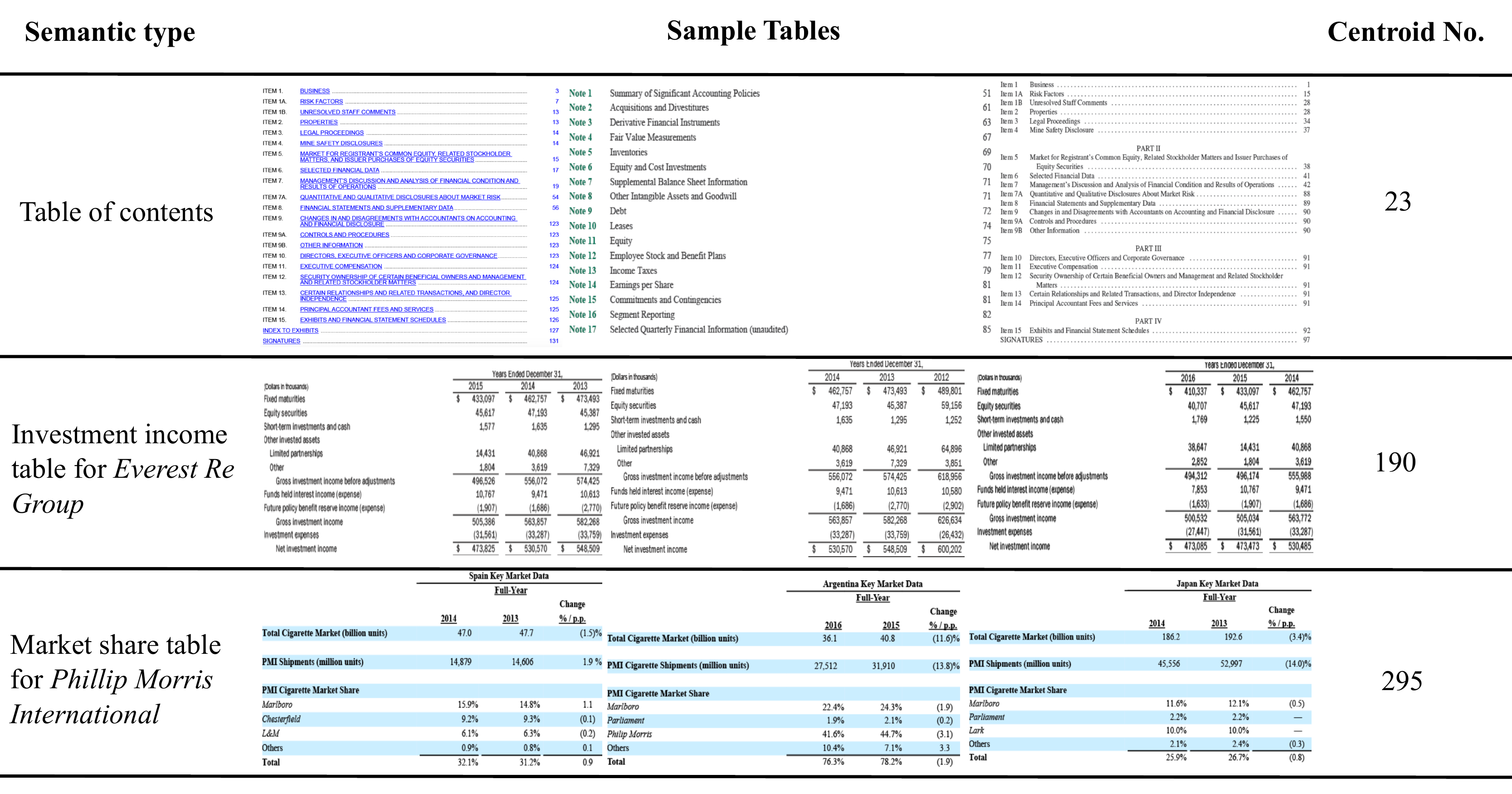}
  \caption{Sample tables from clusters obtained by running $k$-means on \name's \textsc{[cls]} embeddings on the FinTabNet dataset.  \name\ not only clusters embeddings  into reasonable semantic types, such as \emph{Table of Contents} (first row), but it also places tables of the same type from the same company into the same cluster (second and third rows). We provide the source images of the corresponding tables in this figure. }
  \label{fig:clustering}
\end{figure*}

Are the embeddings produced by \name\ useful for clustering and data discovery? To find out, we perform clustering experiments on the FinTabNet dataset from~\citet{Zheng_2021_WACV}. This dataset contains 	$\sim$110K tables from financial reports of corporations in the S\&P-500. We use the \textsc{[cls]} embedding at the $(0, 0)$ position (i.e., the top left-most cell in the table), extracted from a \name\ model trained with the  \freq\ strategy, as a representative embedding for each table in the dataset. Next, we perform $k$-means clustering on these embeddings using the FAISS library \cite{faissJDH17}, with $k$=1024 centroids. While the FinTabNet dataset is restricted to the homogenous domain of financial tables, these tables cluster into sub-types such as \emph{consolidated financial tables}, \emph{jurisdiction tables}, \emph{insurance tables}, etc. We then examine the contents of these clusters (Figure ~\ref{fig:clustering}) and observe that \name\ embeddings can not only be clustered into these sub-types, but also that tables from reports of the same company, but from different financial years, are placed into the same cluster. 
\makeatother

\subsection{Identifying numeric trends}
Next, we analyze how well \name\ understands trends in numerical columns by looking at specific examples of our corrupt cell detection task. The first column of the table in Figure~\ref{fig:num_understand1} contains jersey numbers sorted in ascending order. We swap two cells in this column, \emph{16} and \emph{18}, which violates the increasing trend. Both TaBERT (fine-tuned for corrupt cell detection) and \name\ \freq\ struggle to identify this swap, while \name\ \mix\ is almost certain that the two cells have been corrupted. This qualitative result is further evidence that the \mix\ model has potential for more complex table-based reasoning tasks. 

% \subsection{Nearest Neighbor of Row Embedding}
% \name\ embeds each row as well as its column embeddings. We applied the nearest neighbor to the final layer \textsc{[clsrow]} embedding for visualizing these representations. Fig.\ref{fig:nn_row} is an example result of nearest neighbor for row embedding. We extracted the top two nearest row embedding for the red rows on Fig.\ref{fig:nn_row} (a). Table (b) and (c) are the top two similar embedding results. Though these tables have all different headers, in addition, the values of their leftmost columns are different, the context of these rows are similar. It represents row embedding also learns table context around its row.

\section{Related work}
\label{sec:related_work}
% There are many tasks related to table data, such as column population, row population, corrupted cell classification, column type prediction and table retrieval etc., which are described in the survey paper\cite{10.1145/3372117}. 
% While previous methods to extract table semantics relied on knowledge bases such as DBPedia \cite{10.5555/1785162.1785216} and YAGO \cite{10.1145/1242572.1242667}, our work is part of a recent trend of self-supervised models trained on tabular data.
    The staggering amount of structured relational data in the form of tables on the  Internet has attracted considerable attention from researchers over the past two decades \cite{webtables, limaye-annotating, venetis-recovering, yago, Embley2006TableprocessingPA}, with applications including retrieval \cite{dassarma-finding}, schema-matching \cite{madhavan-schema-2, madhavan-schema-1}, and entity linking \cite{Zhang-entity-linking}. 
    
    Similar to popular large-scale language models pretrained on tasks involving unstructured natural language\cite{Peters:2018,devlin2018bert, Liu2019RoBERTaAR}, our work is part of a recent trend of self-supervised models trained on structured tabular data. TaBERT~\cite{yin20acl} and TaPaS~\cite{Herzig2020TAPASWS} jointly model tables with text (typically captions or questions), and are thus more suited for tasks like question answering  \cite{pasupat2015compositional}. For pretraining, TaBERT attempts to recover the name and data-type of masked column headers (masked column prediction), in addition to contents of a particular cell (cell value recovery). The pretraining objectives of TaPaS, on the other hand, encourage tabular textual entailment. In a concurrent work, the TUTA model~\cite{Wang2020StructureawarePF} uses masked language modeling, cell-level cloze prediction, and table-context retrieval as pretraining objectives. Further, in addition to traditional position embeddings, this work accounts for the hierarchical nature of tabular data using tree-based positional embeddings. Similiarly, in \citet{turl}, the authors perform a variant of MLM called masked entity recovery. In contrast, {\name} is pretrained strictly on tabular data and intended for more general-purpose table-based tasks, and uses corrupt-cell classification as its pretraining task. 
% For column and row population, \cite{entitables} first proposed a knowledge-based solution, while \cite{Zhang:2018:AHT} adapted a word2vec-based method to table headers. This header embedding is only learned from tables. For column type prediction, \cite{Hulsebos2019SherlockAD} proposed a deep learning-based method from single column only, while \cite{zhang2019sato} combined an unsupervised learning method(LDA) with \cite{Hulsebos2019SherlockAD}. 
% In addition, \cite{ijcai2019-289} and \cite{chen2019colnet} are examples of solving column type prediction from table data only.  As stated in the introduction section, these approaches are computationally expensive and they need table-text pairs for pre-training. \cite{deng2020turl} also propose the same architecture as the \cite{Herzig2020TAPASWS} with masked-attention for tiny-BERT network, but \cite{deng2020turl} also use text and table pairs for pre-training, in addition, they learn only entity cells in tables.
\section{Conclusion}
\label{sec:conclusion}

In this paper, we proposed \name, a self-supervised pretraining method for tables without associated text. To reduce the computational cost of training our model, we repurpose the ELECTRA objective for corrupt cell detection, and we use two separate Transformers for rows and columns to minimize complexity associated with sequence length. On three downstream table-based tasks, \name\ achieves competitive or better performance to existing methods such as TaBERT, and an analysis reveals that its representations include a deep semantic understanding of cells, rows, and columns. We publicly release our \name\ pretrained models and code to facilitate future research on tabular representation learning.
% \clearpage
\section{Ethics Statement}
\label{sec:ethics}
As with any research work that involves training large language models, we acknowledge that our work has a negative carbon impact on the environment. A cumulative of 1344 GPU-hours of computation was performed on Tesla V100 GPUs. Total emissions are estimated to be 149.19 kg of CO$_2$ per run of our model (in total, there were two runs). While this is a significant amount (equivalent to $\approx$ 17 gallons of fuel consumed by an average motor vehicle\footnote{\url{https://www.epa.gov/greenvehicles/}}), it is lower than TaBERT’s cost per run by more than a factor of 10 assuming a similar computing platform was used.  
Estimations were conducted using the \href{https://mlco2.github.io/impact#compute}{Machine Learning Impact calculator} presented in \citet{lacoste2019quantifying}.
\section*{Acknowledgements}
\label{sec:acknowledge}

We thank the anonymous reviewers for their useful comments. We thank Christopher Tensmeyer for helpful comments and pointing us to relevant datasets for some of our experiments. We also thank the UMass NLP group for feedback during the paper writing process. This work was made possible by research awards from Sony Corp. and Adobe Inc. MI is also partially supported by  award IIS-1955567 from the National Science
Foundation (NSF).

\bibliography{anthology,bib/naacl2021}
\bibliographystyle{acl_natbib}

\end{document}